\documentclass[journal]{IEEEtran}

\ifCLASSINFOpdf
\else
\fi
\usepackage[linesnumbered, ruled]{algorithm2e}
\usepackage{setspace}

\usepackage{amsmath}
\usepackage{graphicx}
\usepackage{times}
\usepackage{epsfig}
\usepackage{amssymb}
\usepackage{mathrsfs}
\usepackage[table,xcdraw]{xcolor}
\usepackage{color}
\usepackage{url}
\usepackage{siunitx}
\usepackage{booktabs}
\usepackage{tabularx}
\usepackage{float}
\usepackage{threeparttable}

\ifCLASSINFOpdf

\else

\fi

\begin{document}
	\title{Minimum Potential Energy of Point Cloud for Robust Global Registration}
	
	\author{Zijie Wu,~Yaonan Wang,~Qing Zhu,~Jianxu Mao,~Haotian Wu,~Mingtao Feng and Ajmal Mian		
		\thanks{Z. Wu, Y. Wang, Q. Zhu, J. Mao and H. Wu are with the College of Electrical and Information Engineering, Hunan University, Changsha 410082, China (email:wuzijieeee,zhuqing,maojianxu,wuhaotian,yaonan@hnu.edu.cn).}
		
		\thanks{M. Feng is with the School of Computer Science and Technology, Xidian University, Xi’an 710071, China (email: mintfeng@hnu.edu.cn).}
		
		\thanks{A. Mian is with the Department of Computer Science and Software Engineering, The University of Western Australia, Perth, Crawley, WA 6009, Australia (e-mail: ajmal.mian@uwa.edu.au).}
		
		\thanks{Manuscript received ***; revised ***.}}
	
	\maketitle
	
	\begin{abstract}
		In this paper, we propose a novel minimum gravitational potential energy (MPE)-based algorithm for global point set registration. The feature descriptors extraction algorithms have emerged as the standard approach to align point sets in the past few decades. However, the alignment can be challenging to take effect when the point set suffers from raw point data problems such as noises (Gaussian and Uniformly). Different from the most existing point set registration methods which usually extract the descriptors to find correspondences between point sets, our proposed MPE alignment method is able to handle large scale raw data offset without depending on traditional descriptors extraction, whether for the local or global registration methods. We decompose the solution into a global optimal convex approximation and the fast descent process to a local minimum. For the approximation step, the proposed minimum potential energy (MPE) approach consists of two main steps. Firstly, according to the construction of the force traction operator, we could simply compute the position of the potential energy minimum; Secondly, with respect to the finding of the MPE point, we propose a new theory that employs the two flags to observe the status of the registration procedure. The method of fast descent process to the minimum that we employed is the iterative closest point algorithm; it can achieve the global minimum. We demonstrate the performance of the proposed algorithm on synthetic data as well as on real data. The proposed method outperforms the other global methods in terms of both efficiency, accuracy and noise resistance.
	\end{abstract}
	
	\begin{IEEEkeywords}
		Point set registration, force traction, global robust method
	\end{IEEEkeywords}

	%
	\IEEEpeerreviewmaketitle

	\section{Introduction}
	\IEEEPARstart{P}{oint} cloud registration (or alignment) is a fundamental problem in computer vision and robotics. It is the task of establishing correspondences between a pair of point clouds, each residing in a different coordinate system, and subsequently minimizing the distances between the corresponding point pairs to register the two point clouds in the same coordinate system. Many application domains rely on point set alignments, such as 3D reconstruction, shape recognition, map relocalization, computer-aided medical diagnosis, and so on. In recent years, the ubiquity of 3D acquisition devices has lead to a growing interest in point cloud registration and the need for more effective, robust and efficient algorithms. However, 3D acquisition devices, especially the ones that operate in realtime, provide noisy point cloud data which makes it challenging to achieve efficient and accurate registration.

	Over the past few decades, numerous methods have been developed for point cloud registration. The Iterative Closest Point (ICP) algorithm~\cite{icp1,icp2} has been widely used for rigid registration of 3D point clouds due to its simplicity and performance. In its basic form, ICP first establishes correspondences between the nearest points of the two point clouds and then minimized the $\ell_2$ distance between the corresponding point pairs. Hence, ICP assumes that the nearest points of the two point clouds are corresponding points. This  assumption means that the two point clouds must already be approximately registered. ICP computes the rotation and translation that would minimize the $\ell_2$ distance between the corresponding points. It then applies the rotation and translation to one of point clouds and repeats the process i.e. establishes correspondence between nearest points which may be different now. It iterates until the rotation and translation converges to zero. The ICP algorithm is intuitive and easy to implement in practice because of its conceptual simplicity.

	However, the ICP algorithm is well-known to be susceptible to the local minima problem due to its assumption that the set of nearest points in the current iteration will be better correspondences than those in the last iteration.
	This assumption can easily fail when the point cloud data is contaminated with noise or has missing regions of overlap. 
	
	The non-convex nature of the registration problem makes it inherently vulnerable to the local minima problem and sensitive to satisfactory initialization on a case by case basis depending on the shape represented by the point clouds. As such, the ICP method does not guarantee a globally optimal registration and there is no effective way to automatically determine if it is trapped in a local minimum that is too far from the global optimum. 
	
	To deal with the local minima problem, Chetverikov \textit{et al.} proposed the trimmed ICP algorithm~\cite{Trimmedicp} which allows the application of ICP to point clouds with partial overlap. 
	Fitzgibbon \textit{et al.}~\cite{OT1} employed a standard iterative non-linear optimizer (LM algorithm) to replace the closed-form $\ell_2$ minimization part of the ICP.
	Nonetheless, the original ICP and all its variants are still sensitive to noise and necessitate good initialization of the registration to start with. Few heuristic methods~\cite{pf}~\cite{sa} have also been presented to alleviate the local minima problem. Another strategy~\cite{lei2017fast} is to use coarse alignment with other methods, such as feature matching, to achieve a good initialization. However, feature-based methods are not always reliable and do not guarantee a globally optimal transformation. More critically, due to the $\ell_2$-norm least squares~\cite{mactavish2015all}, when minimization is applied, the optimizer runs even after reaching the global minimum, especially when dealing with contaminated real point cloud data. A small number of outliers may adversely affect the validity of the results. There are several methods proposed to deal with outliers~\cite{Trimmedicp}~\cite{OT1}~\cite{OT2}~\cite{OT3} and for computing consensus maximization (point pairs matching maximum) between the point clouds~\cite{cmax1}~\cite{cmax2}~\cite{cmax3}. However, these methods are based on heuristics and incur additional computational costs.

	We propose a robust global point cloud registration algorithm that does not require point correspondences. We introduce physics in an intuitive way to solve this computer vision problem to achieve very efficient registration. Similar to the feature-based strategy, we also essentially employ a coarse alignment to fulfil a good initialization. However, in contrast, our coarse alignment method is much more efficient and obviates the drawbacks of conventional methods. We coin our method as the Minimum Potential Energy (MPE) of point clouds.
	
	We also extend our method to work at larger scales and present the MPL algorithm based on the proposed MPE theory for more efficient registration. Specifically, we propose the idea of $P_2$ least criterion, allowing us to judge exactly whether the iteration is a globally optimal registration to facilitate the computation of the point cloud coarse alignment with torque and vector in the physics system. Finally, we employ a trimmed ICP method to precisely and efficiently approach the globally optimal registration. The method that we propose is optimal, guaranteed by the shape of the objects which are different from extracting a 3D convex hull~\cite{3points}. The shape influenced by multi-points of inliers decide the results of the optimal registration. 
	
	We validate the proposed method with a standard point cloud repository and compare it to several existing methods. Our method is able to register the real urban dataset and shows exceptional robustness to noise and outliers. We also report detailed analysis of our method in terms of time complexity and accuracy. Moreover, compared to  traditional descriptor-based methods such as FPFH~\cite{FPFH} etc., the proposed method is more efficient in dealing with point clouds of surfaces that have small variations in geometric shape or in the curvature.

	%
	%

	\begin{figure*}[t!] 
		\center{\includegraphics[width=0.99\textwidth]{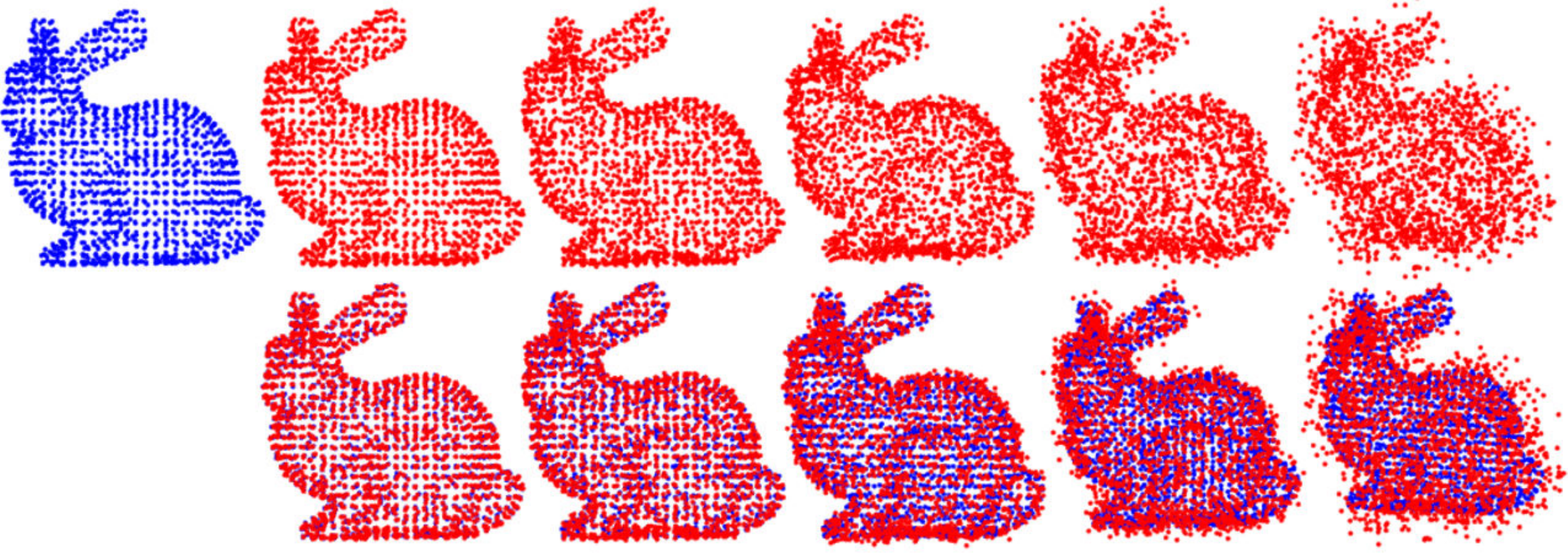}}
		\caption{\label{fig:gaussian_demo} Top row: Five template point clouds (red) are generated from a reference point cloud (blue) with varying scales of Gaussian noise. Bottom row: The reference point cloud has been correctly registered to all templates.}
	\end{figure*}
	\section{Related Work}
	A large volume of work has been published on point cloud registration. A complete list is beyond the scope of this paper. However, to put our proposed method in context, we provide a brief review of local and global registration methods. 
	
	\subsection{Local Methods}
	The ICP algorithm~\cite{icp1}~\cite{icp2} is the most popular method for point cloud registration in the local category. Its introduction serves as an important milestone for the this line of research. Despite its many desirable properties, including simplicity, ICP implicitly requires a full overlap between the point clouds, which is a rare situation in practice. Therefore, the trimmed ICP algorithm~\cite{Trimmedicp} was proposed by Chetverikov \textit{et al.} as an extension to ICP for application to asymmetric point clouds. Fitzgibbon \textit{et al.}~\cite{OT1} optimized the point cloud registration function of the Levenberg-Marquardt algorithm and provided better convergence than ICP. However, without good initialization, the ICP and all its variants have high chances of getting trapped in a local minimum.
	
	Another line of point cloud registration methods include the adoption of probabilistic models. Jian~\textit{et al.}~\cite{GMM1}~\cite{GMM2} proposed a more robust method by representing a point cloud with a Gaussian mixture model (GMM) and formulated the ICP algorithm as a special case of minimizing the Kullback-Leibler divergence between two GMMs. Granger~\textit{et al.}~\cite{EMICP} introduced the expectation-maximization (EM) algorithm for registration error optimization, achieved by the probabilities of computation of the E-step and the parameters update of the M-step. The normal distributions transform the (NDT)~\cite{NDT1}~\cite{NDT2} algorithm and performs the optimization using the Gaussian models' definition of each cell in a spatial grid. Magnusson \textit{et al.}~\cite{NDT3} experimentally showed more robust performance than the ICP in handling large transformations.
	Campbell~\textit{et al.}~\cite{SVR} proposed to construct the GMM by using support vector machines. Despite the fact that the GMM-based methods~\cite{GMM1}~\cite{GMM2}~\cite{KC} show more robust registration performance with poor initialization, all these methods are still rely on local search and can still get stuck in a local minimum.

	\subsection{Global Methods}
	To address the local minima problem, a large number of global methods have been proposed. Typical methods include genetic algorithms~\cite{ge1}~\cite{ge2}, simulated annealing~\cite{sa}, particle filtering~\cite{pf} etc., which apply stochastic optimization to avoid local minima. These methods have alleviated the local minima problem to great extent. However, these methods still require good initialization, otherwise they must explore a large parameter space using heuristic search methods. Moreover, they still do not guarantee a globally optimal solution.

	Descriptor-based methods, such as FPFH~\cite{FPFH}, key-point detection~\cite{mian2010repeatability}, and fast descriptor~\cite{lei2017fast}, extract local feature descriptors at some key-points. These descriptors of two two point clouds are then matched to find corresponding points. A transformation is then computed based on the corresponding points and applied to one of the point clouds for coarse alignment. This alignment is coarse due to errors in the corresponding points, however, it still serves as a good initialization to find the global optimum. Aiger~\textit{et al.} proposed the 4PCS method that could efficiently extract coplanar four points from the point cloud and establish the congruent coplanar four-points sets correspondences. The process of a RANSAC-based algorithm~\cite{ransac} also brings in the quadratic time complexity. Mellado~\textit{et al.}~\cite{S4PCS} proposed the super-4PCS method for reducing the time complexity to linear. Lei~\textit{et al.} employed the congruent differential information to search for the best correspondence and fulfilled a great coarse registration initialization efficiently. These descriptor-based methods are often effective on surfaces where the descriptors can be computed repeatability of the keypoints and uniqueness of the descriptors. These conditions are not always met in high noise levels and weak shape (texture) variations. 
	
	Some solutions aim to achieve the global optimal registration by introducing a branch and bound (BnB) approach that divides the parameter space into multiple sub-spaces and compute the bounds iteratively. Yang~\textit{et al.}~\cite{go-icp} proposed the Go-ICP algorithm that combined the BnB (Branch and Bound) and ICP to approach the global minimum. This is the first effective global registration method in practice. Many follow up BnB-based methods have been proposed since then. Some of them focus on tighter bounds~\cite{tb1}~\cite{tb2}~\cite{tb3}~\cite{tb4}~\cite{go-icp}, while others~\cite{speed1} aim to speed-up the search itself. Although these methods aim to provide a globally optimal guarantee, all of them run, in the worst case, with exponential time complexities. Golyanik~\textit{et al.}~\cite{gaa} proposed a completely different method that uses the gravitational pull between point clouds to register them. This method formulates point cloud registration as a gravitation problem and showed good performance. However, it is still unable to approach a globally optimal registration especially in the presence of noise and outliers. Moreover, the method is extremely inefficient, requires a full-overlap between the two point clouds and cannot deal with problems such as hollow point clouds.

	\begin{table}[htbp]
		\centering
		\caption{\label{tab:ET} Registration results on Bunny dataset with increasing down-sampling ratio. Evaluation metric for error is $L_2$ distance.}
		\begin{tabular}{p{30pt}<{\centering}p{60pt}<{\centering}p{50pt}<{\centering}p{65pt}<{\centering}}
			\toprule
			Sampling Ratio & Average Run Time & Estimation Error & Estimation Error variance\\
			\midrule
			0.05 	& 0.705879 &0.010845 & 3.21E-05\\
			0.10	& 2.889829 &0.008355 & 2.07E-05\\ 
			\textbf{0.15}	& \textbf{6.223227} &\textbf{0.004785} & \textbf{8.99E-06}\\
			0.20	& 11.26921 &0.007065 & 1.63E-05\\
			\textbf{0.25}	& \textbf{18.46324} &\textbf{0.005311} & \textbf{3.67E-06}\\
			0.30	& 25.43114 &0.004270 & 3.03E-06\\
			0.35	& 36.03378 &0.003597 & 4.85E-06\\
			0.40	& 47.33147 &0.002728 & 2.75E-06\\
			0.45	& 60.27947 &0.003204 & 2.18E-06\\
			0.50	& 255.3882 &0.002947 & 2.07E-06\\
			0.55	& 95.16613 &0.002630 & 1.07E-06\\
			0.60	& 115.0996 &0.002449 & 7.83E-07\\
			0.65	& 133.8049 &0.001972 & 9.74E-07\\
			0.70	& 148.7615 &0.001974 & 1.13E-06\\
			0.75	& 177.2722 &0.001821 & 4.90E-07\\
			0.80	& 194.7487 &0.002077 & 6.49E-07\\
			\bottomrule
		\end{tabular}
	\end{table}

	\section{Approach}
	Point cloud registration problem generally involves two D-dimensional point sets, a reference set and a template set. The template set is rotated and translated so that it aligns with the reference set. Let $\mathbf{X}_{D\times N}=\begin{Bmatrix}x_1,x_2,......x_N\end{Bmatrix}$ represent the template set and  $\mathbf{Y}_{D\times M}=\begin{Bmatrix}y_1,y_2,......y_M\end{Bmatrix}$ the reference set. 
	We introduce a physical system into this problem to facilitate efficient registration. The system is adapted from the real world by adding a series of modifications to adjust it to the problem at hand. These modifications and assumptions are presented below:
	
	\begin{enumerate}
		\item Every point is assumed to be a particle that has a mass but no volume to avoid collision issues in the system.
		\item The point cloud $X$ is considered to be a rigid body. The force between its internal points belongs to the internal force of the system and does not affect its motion.
		\item The point cloud $X$ resides in the constant in-homogeneous force field induced by the point cloud $Y$.
		\item The system does not follow the law of conservation of kinetic energy, meaning that the system is not isolated. Moreover, each position of $X$ is computed and its distribution is discrete.
		\item Potential energy at infinity is regarded to be zero.
	\end{enumerate}
	
	Before introducing the proposed method, we first review two highly relevant methods, the ICP method~\ref{iiia} and the N-body simulation~\ref{iiib}. We formulate the point cloud registration problem in Section~\ref{iiia}, then perform the MPE problem reformulation in Section~\ref{iiic}. Besides, we also introduce the force traction computed in Section~\ref{iiid} and discuss the global minimum approximation theory in Section~\ref{iiie}.

	\subsection{\label{iiia} Problem Formulation}
	Point set registration is the problem of finding the best transformation parameters between $K$ sets where $K=2$ in this paper. The best transformation can be expressed as a tuple $(R,t)$. For $x_i\subset\mathbf{X}$ and $y_i\subset\mathbf{Y}$, the former can be aligned to the latter as:
	\begin{equation}\label{eq1}y_j=Rx_i+t+o_i+\epsilon_i, \end{equation} 
	where $R\in SO(3)$, and $t\in {R}^3$ are unknown rotation matrix and translation vector respectively. The $o_i$ contains an arbitrary number of outliers and  $\epsilon_i$ represents the sampling noise. 
	When multiple pairs of inliers exist,  the correspondence tuple $(R,t)$ are constructed between $(x_i,y_i)$. While the point pair $(x_i,y_i)$ do not belong to the existing inlier collection, the $o_i$ is an arbitrary non-zero vector. 
	Given a point pair, whether or not it is an inlier must be judged by the algorithm. The sampling noise $\epsilon_i$ is intrinsic to the 3D scanning process and hence unavoidable. For the above formulation, we apply the $L_2$ distance evaluation equation:\begin{equation}\label{eq2} E(R,t)=\sum_{i=1}^{N}e_i(R,t)^2=\sum_{i=1}^{N} (\left \| y_j-(Rx_i+t) \right \|)^2, \end{equation} 
	where, $y_j$ denotes the optimal correspondence point for $x_i$ in \textbf{Y}, and $e_i(R,t)$ represents the i-th residual error for $x_i$. To achieve the optimal registration, the parameter $o_i$ should be a zero vector, and the vector $\epsilon_i$ should be minimized.
	
	The ICP algorithm performs optimization by alternately applying the following two functions:
	\begin{equation}\label{eq3} E(R,t)=\sum_{i=1}^{N}e_i(R,t)^2=\sum_{i=1}^{N} \left \| y_{j^*}-(Rx_i+t) \right \|^2, \end{equation} 
	\begin{equation}\label{eq4} j^*=\arg\min  \left \| y_j-(Rx_i+t) \right \|,\end{equation}
	where $y_{j^*}$ and $x_i$ denote the optimal corresponding point pair. Equation~\ref{eq3} implements the transformation estimation and equation~\ref{eq4} matches the closest points.
	
	Taking nearest neighbor points as the corresponding points makes the ICP algorithm susceptible to getting trapped in local minima. We argue that the matching/corresponding points of a global optimization algorithm should be computed globally. To achieve this, we propose the MPE method which considers all points during matching while distributing their weights.

	\subsection{\label{iiib} N-Body Simulation}
	The	N-body simulation~\cite{NBS} involves formulating the dynamics of a large number of mutually attractive particles to model the astrophysical phenomena. It covers many applications for the problem of star-cluster dynamics, globular clusters, etc. The interactions between celestial bodies are governed by Newton's law of gravity in a superimposed gravitational field. In the classical N-body problem, the celestial object contains a set of particles $i \in \{1,2,...,N\} $ and induces a constant inhomogeneous gravitational field in a system taking the following form:
	\begin{equation}\label{eqnb}
	\textbf{F}_{i}=-Gm_{i}\sum_{j}^{N}\frac{m_{j}(\textbf{r}_{i}-\textbf{r}_{j})}{\left \| \textbf{r}_{i}-\textbf{r}_{j} \right \|^3},
	\end{equation}
	where $\textbf{F}_{i}$ is the force of the i-th celestial body and $G$ is the  gravitational proportionality constant connecting the bodies in Newton's law. $m_{i},m_{j} $ denote the particle masses and $\textbf{r}_{i},\textbf{r}_{j}$ represent the position vectors of celestial bodies $i,j$, respectively. $ \left \| \cdot \right \| $ denotes the $L_2$-norm distance. 
	
	To avoid problems, such as turning the normal aster into a black-hole celestial body, the assumptions and modifications are refined by setting a upper bound of traction. 
	In this paper, we focus on the collisionless N-body simulation i.e. merging, splitting and masses transfer are not taken into account.
	
	In the N-body problem, the total energy of the i-th particle $ \phi_{x_i} $ is equal to the sum of the negative gravitational potential energy $ \phi_{i}^{a}  $, kinetic energy ($K_E$) and some external potential $ \phi_{i}^{ext} $. By convention, the reference location with zero gravitational potential energy is infinitely far away form any particle with mass, resulting in negative potential energy in any location of the system:
	\begin{equation}\label{eqnb2}
	\phi_{x_i}=-\phi_{i}^{a}-\phi_{i}^{ext}+K_E
	\end{equation}
	Without considering external forces, any energy dissipation can reduce the total energy to the minimum potential energy (MPE) i.e. $K_E=0$ in this state.

	\subsection{\label{iiic} MPE Reformulation}
	To overcome the local minimum problem, a global registration method must have two properties. One, the local matching process (e.g. based on nearest points) must be replaced with a process that is influenced globally. 
	Two, different weights should be allocated to the points to alleviate the influence of outliers. 
	We first propose MPE, a physics inspired procedure based on the modified N-body simulation, to search for fine registration. Unlike ICP, which relies on closest point-pairs, we use the gravitational field induced by the set of points (particles) to implement the registration.

	\begin{figure}[t!] 
		\center{\includegraphics[width=0.5\textwidth]{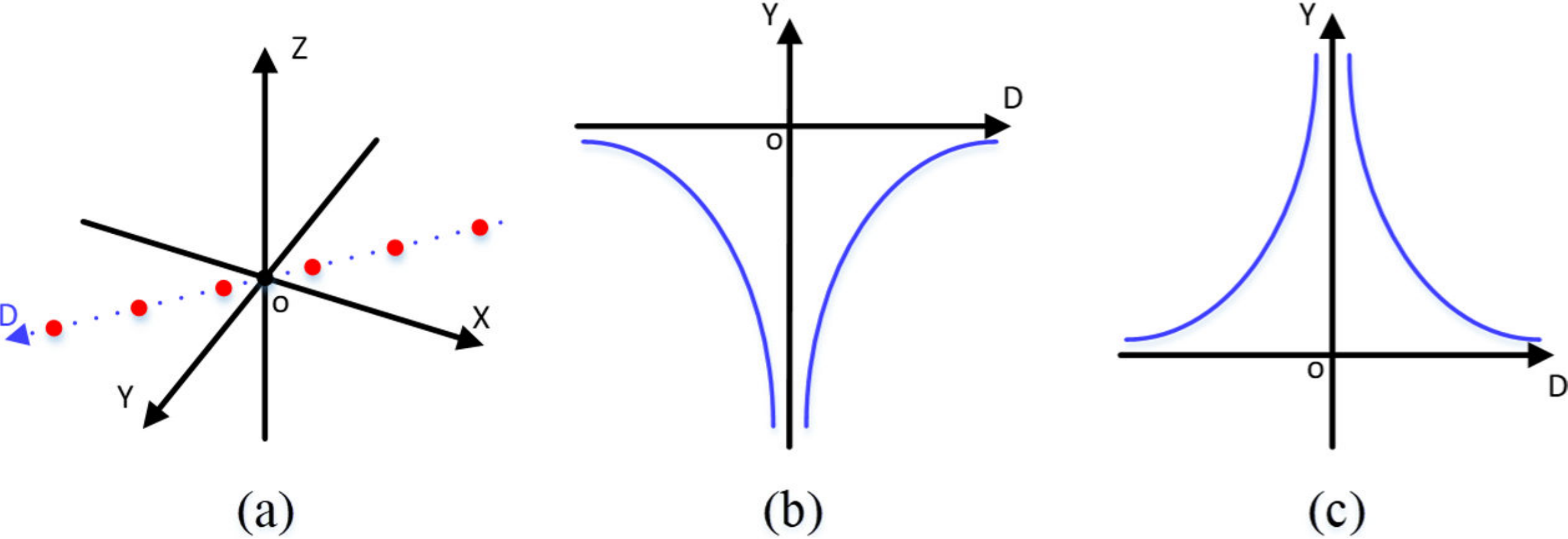}}
		\vspace{-5mm}
		\caption{\label{fig:crosspoint} Illustration of potential energy and gravitation cross functions with respect to linear motion. (a) A red point moves along the axis D in 3D space. (b) The potential energy of the red point as it travels along the axis D. (c) The gravitation on the red point induced by particle o as the point moves along the axis D.  
		}
	\end{figure}

	The physics-based method introduced above, incorporates gravitation into point cloud registration. To compare the physical parameters with the registration parameters; the physical position minimum $E(R,t)$~\cite{miniPE} is the 
	best registration position. Figure~\ref{fig:crosspoint} shows the cross function distribution schematic diagram of a single pair of points. The PE of the whole system is formulated as:
	\begin{equation}
	\label{eq5}-\phi_{i}^{a}=\sum_{i}^{N}\sum_{j}^{M} \int -\frac{GMm}{R^3}\cdot \vec{R} dr \end{equation} 
	Where, $R$ denotes the Euclidean distance between points $(i,j)$ and $\vec{R}$ represents the related vector. The PE of the system follows the fifth assumption i.e. it is zero at infinity. Take a point pair $(x_i,y_j)$ as an example and put $y_j$ in place of point o in Figure~\ref{fig:crosspoint}. Figure~\ref{fig:crosspoint}(a) shows $x_i$ (red point) traveling in space through $y_j$ (original point) along a straight line D, where D as a single one-dimensional transverse axis. Figure~\ref{fig:crosspoint}(b)(c) show the function of PE and the traction force as the distance between the point pair $(x_i,y_j)$ changes. Note that the limits of the PE and traction force are crossed simultaneously. When the Euclidean distance $R$ approaches zero, the PE approaches infinity and at this moment, the point pair $(x_i,y_j)$ is also registered.
	
	In terms of capturing the mutation of function~\ref{eq5}, we can easily approach the mutation point using our method. After the problem reformulation, the original point cloud registration problem is transformed into parameter searching of the traction force mutation point. To summarize, the optimal registration point pairs induce the minimum potential energy of the two point clouds system. The minimum potential energy point pairs are the best correspondence and their sets consist the optimal registration of the two point clouds.

	\subsection{\label{iiid} $P_2$ Criterion and Algorithm Implementation}
	It is easy to see from (\ref{eq3}) that most current registrations are optimized by minimizing the $L_2$ distance function $E(R,t)$. However, it is well-known that the $L_2$-norm least squares is not robust as it is sensitive to outliers. To make matters worse in the specific case of point set registration based on $L_2$-norm optimization, incorrect matching pairs receive more weight given that their Euclidean distance is very high. Such pairs should be assigned lower weights.
	
	To address the above problem, based on the MPE method of the N-body simulation, we propose a new $P_2$ evaluation that distributes appropriate weight as the Euclidean distance rises and minimizes the following $P_2$ error
	\begin{equation}
	\label{eq33} E_p(R,t)=\sum_{i=1}^{N}e_{pi}(R,t)
	=-\sum_{i=1}^{N} \frac{K}{\left \| y_{j^*}-(Rx_i+t) \right \|^2 + \varepsilon^2}, 
	\end{equation} 
	where $e_{pi}(R,t)$ denotes the per-point residual error of $x_i$, and $y_{j^*}$ and $x_i$ represent the optimal correspondence point pair. Given $R\in SO(3)$ and $t\in {R}^3$, the $P_2$ criterion allows outliers in data set, and hence improves the robustness of the proposed registration method. $\varepsilon^2$ guarantees a lower bound single point pair, precluding the registration from the incalculable infinitesimal singularity. K is a user defined constant which adjusts the weight of the point cloud and can be adaptive with additional prior information such as density and curvature.
	
\begin{figure}[t!] 
		\center{\includegraphics[width=0.5\textwidth]{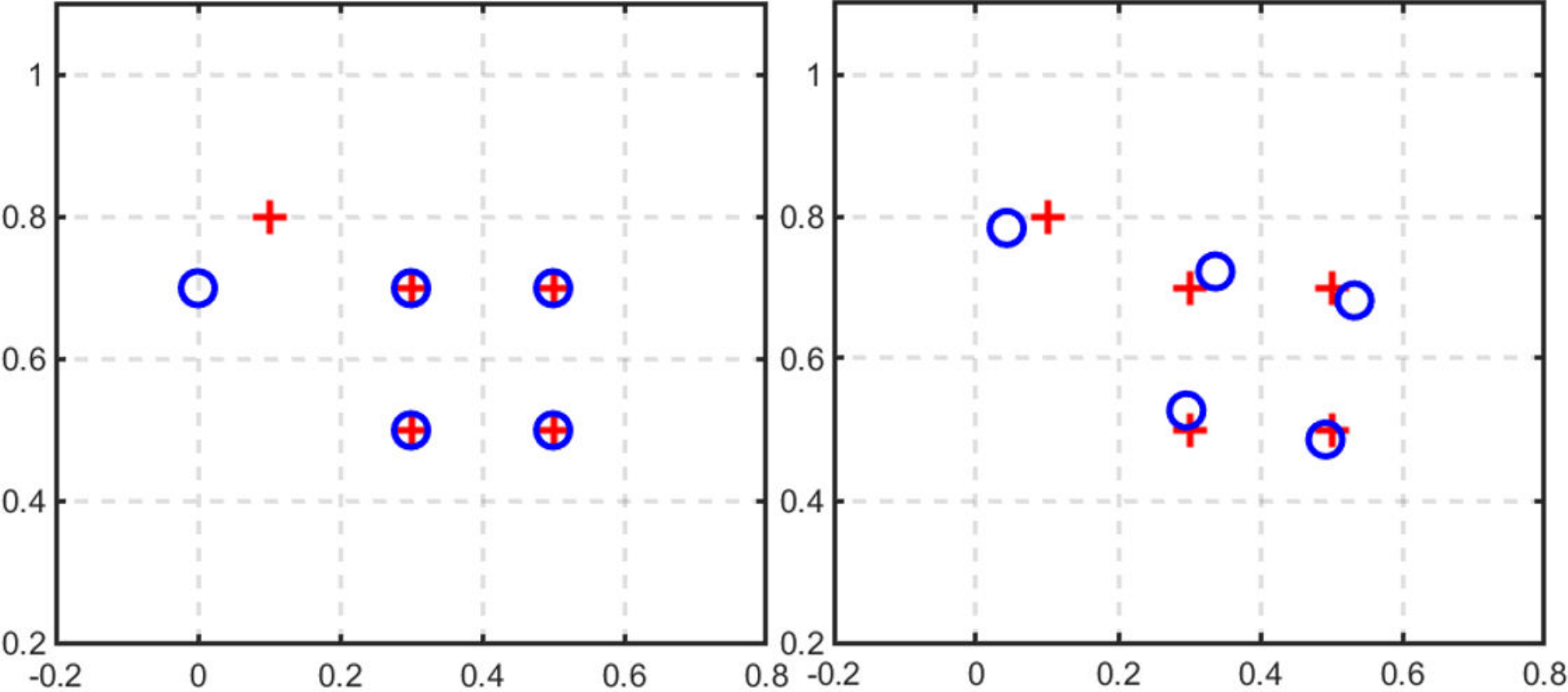}}
		\vspace{-5mm}
		\caption{\label{fig:P2} 2D illustration of the difference between $P_2$ and $L_2$ criteria. Two point sets (blue and red) with one outlier pair are to be registered. Notice that the $P_2$ registration (left) is not effected by the outlier pair of points whereas the $L_2$ registration (right) is effected. 
		} 
	\end{figure}

	The traditional $L_2$ distance method in ICP uses a non-linear loss function $ min(\left \| y_{i}-(Rx_i+t) \right \|^2) $ of the point-pairs to distribute weights at different distances. This loss function generally works but fails in the presence of noise and outliers. Unlike our method, its quadratic weight biases the registration. Our proposed $P_2$ criterion easily ignores the noises and distributes lower weights to points that are far.
	Figure~\ref{fig:P2} shows a toy 2D example to illustrate the difference in alignment that results from the $P_2$ and $L_2$ criterion. The left image is the registration result employing $P_2$ minimization, and the alignment procedure stops when the global optimal is approached. The right image denotes the $L_2$-norm least square minimization where the pair of outliers lead to an erroneous registration. $P_2$ optimization is sensitive to changes of point pair residual error and reduce $e_{pi}(R,t)$ as the point pair error rises (zero in infinite distance).

	\begin{algorithm}
		\label{algorithm1}
		\SetAlgoLined
		\KwIn{Point Cloud P, Q}
		\KwOut{Rotation matrix \textbf{R}, translation vector \textbf{t}}
		init flags $F_R = \vec{0}$, $F_t = \vec{0}$ and steps $s_R$, $s_t$;\\
		\While{$s_R>\varepsilon_R$ AND $s_t>\varepsilon_t$}{
			$F_{R_{last}}=F_R$, $F_{t_{last}}=F_t$;\\
			compute $F_R$, $F_t$;\\
			\If{$F_{R_{last}}\cdot F_R <0$}{
				$s_R=\frac{s_R}{2}$;\\
				\If{$F_{t_{last}}\cdot F_t <0$}{
					$s_t=\frac{s_t}{2}$;\\
				}
			}
			Transform Q with $s_R, s_t$;	
		}
		\caption{Minimum Potential Energy (MPE) Algorithm for traction force to optimal registration}
	\end{algorithm}

	\begin{figure}[t!] 
		\center{\includegraphics[width=0.5\textwidth]{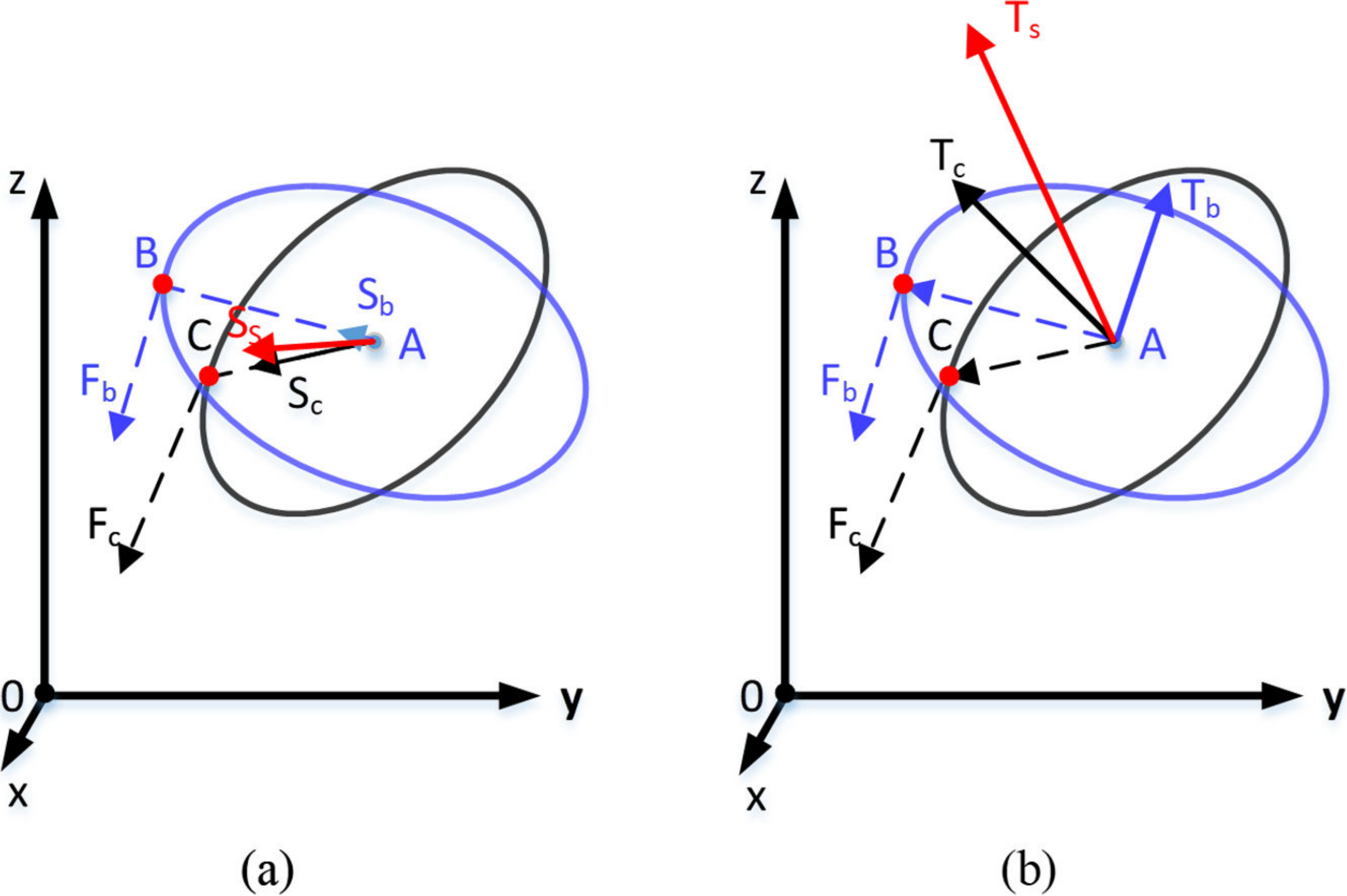}}
		\vspace{-5mm}
		\caption{\label{fig:MPE} Illustration of MPE the rotational torque $T_s$ and gravitational vector $S_s$ computation in 3D Cartesian coordinate system. Points A, B, C are in a same point sets and A is specified as the center.
			The blue ellipse denote a circle with the length of AB as the radius and A as the center in 3D coordinate, same as the black circle. $BF_b$, $CF_c$ are the gravities and direction from B and C to original point o respectively. (a) The gravitational vector computation, composed by all gravitational component (for instance, $S_b$, the component of force $BF_b$ in the direction of AB). (b) The rotation torque computation, composed by all rotation torque component (for instance, $T_b$, computed by the tangential component of the $BF_b$ and perform a cross-product with vector AB).}
	\end{figure}

	Algorithm \ref{algorithm1} details the steps of the proposed MPE registration. 
	From Section \ref{iiic}, we converted the conventional point cloud registration problem to the zero-point saltus step parameter search for the traction force. The traction force is defined as
	\begin{equation}
	\label{eq6} 
	\textbf{F}_{xi}=-Gm_{xi}\sum_{j}^{M}\frac{m_{yj}}{\left \| c_{yj}-c_{xi} \right \|^2}\cdot\vec{n_{ij}},
	\end{equation}
	where, $m_{xi}$ and $m_{yj}$ denote the masses of $i$-th and $j$-th points from the related $X$ and $Y$ point clouds respectively. The symbols $c_{xi}$ and $c_{yj}$ are the absolute coordinates of the points $X_i$ and $Y_j$, respectively. Moreover, $\vec{n_{ij}}=\frac{c_{xi}-c_{yj}}{\left \| c_{xi}-c_{yj} \right \|}$ is the normalized vector that point particle $x_i$ to particle $y_j$. Every single point is regarded as a particle i.e. each point has ideal conditions-having mass but no volume or shape. Using $R_{ij}$ as an abbreviation representing $(c_{xi}-c_{yj})$, $R_{ij}$ produces the infinite value of the traction force ${F}_{xi}$. However, if the distance between the two points $R_{ij}$ becomes infinitesimally small, it will trap the entire point cloud in a singularity, akin to a black hole. To avoid this situation, we introduced a non-zero real number $\varepsilon $ that produces a minimum traction force to realize an extreme accumulative value. The modified equation is then  
	\begin{equation}\label{eq7} 
	\textbf{F}_{xi}=-\sum_{j}^{M}\frac{K_i}{\left \| R_{ij} \right \|^2+\varepsilon^2}\cdot\vec{n_{ij}}, \end{equation}
	which allows us to promote the whole point cloud from a single pair of point sets. $ K_i $ is a hyper-parameter to control the point cloud density distribution. For capturing the mutation, a pair of flags is set, $F_R$ and $F_t$, to be read to observe the changes of rotational torque and the gravitational vector, respectively. Furthermore, to simplify calculation, we take a point C (could be an arbitrary point but usually is the center of the point cloud) and use $\vec{v_i}=\vec{(c_{xi}-c)}$ to represent the unit vector direction from V to point $x_i$. The choice of C influences the time of convergence. The rotational torque and gravitational vector are computed as: 
	\begin{equation}\label{eq8} T=\sum_{i}^{M}\vec{v_i}\times \begin{bmatrix}
	\textbf{F}_{xi}-(\textbf{F}_{xi}\cdot \vec{v_i})\vec{v_i} \end{bmatrix}\end{equation}
	\begin{equation}\label{eq9} S=\sum_{i}^{M}(\textbf{F}_{xi}\cdot \vec{v_i})\vec{v_i}\end{equation}
	Where, $T$ denotes the rotational torque and $S$ represents the gravitational vector. In each iteration, we recompute the $(T,S)$ and store its last values. The flags are defined as follows: $F_R=\textbf{T}*\textbf{T}_{last}$, $F_t=\textbf{S}*\textbf{S}_{last}$. When the flag values are less than zero, we assume that the point cloud has crossed the mutation point i.e. the force direction has reversed.

	Figure \ref{fig:MPE} illustrates the derivation procedure using only a dual point system for ease of understanding. The figure shows the rotational torque and the gravitational vector computation of point B and C, under the gravitational field of point O. Point A is picked to be the computation center (the position of center C discussed above). There are two planes, defined by points ABO (the blue circle) and point ACO (the gray circle). $F_b$ and $F_c$ are the gravities generated by the traction from O to B and C. $S_b$ and $S_c$ denote the component of $F_b$ and $F_c$ in the axial direction that are then combined into the total gravitational vector $S_s$. The rotational torque $T_s$ is computed in a similar way.
	
	\begin{figure}[t!] 
		\center{\includegraphics[width=0.5\textwidth]{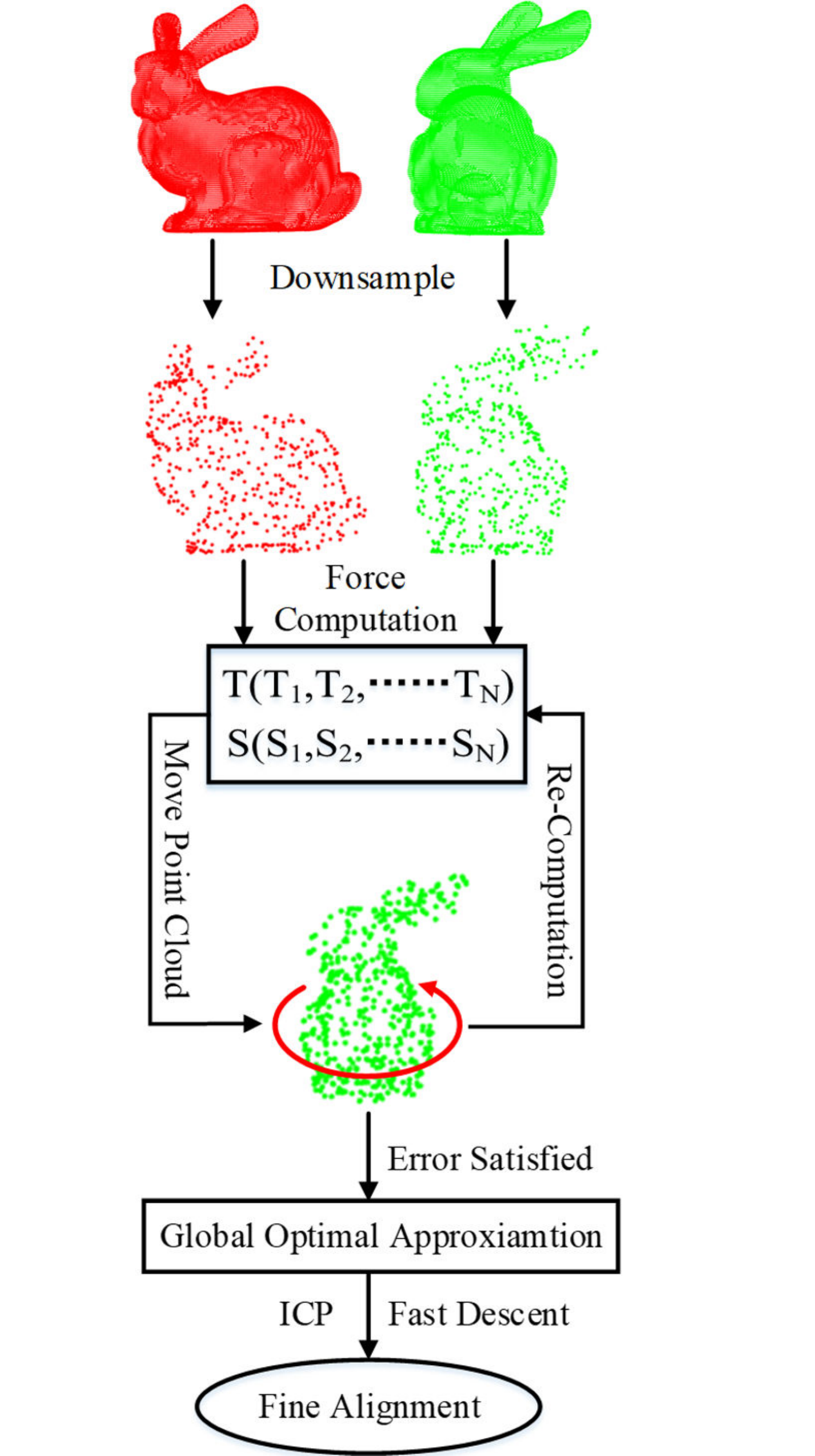}}
		\vspace{-5mm}
		\caption{\label{fig:MPL} Flow chart of our proposed algorithm (MPL). The vectors T and S denote the rotation torque and the gravitational vector respectively. When the condition for breaking the loop is satisfied, we employ the ICP algorithm to achieve fast descent to the global optimal minimum.}
	\end{figure}

	\subsection{\label{iiie} Global Minimum Approximation Theory}
	The proposed MPE is a global registration method. In Section \ref{iiic}, we showed that the MPE mutation point reaches the position of the best point cloud registration. A limitation of MPE is that its time complexity exponentially rises with the number of points in the point cloud. In this Section, we extend the MPE theory and propose MPL to handle larger scale point clouds.
	
	For scalability to large point clouds, we propose a strategy for down point cloud sampling. One of the kernels for our point cloud registration method is the center of mass. Statistically, random down sampling does not change the gravitational field distribution of the raw point cloud and still reduces time complexity. We down sample the raw point sets to get the sparse sets $\mathbf{X}_{d}=\begin{Bmatrix}x_1,x_2,......x_N\end{Bmatrix}$ and $\mathbf{Y}_{d}=\begin{Bmatrix}y_1,y_2,......y_M\end{Bmatrix}$. The new sparse point cloud provides the same distribution as the original one but with smaller intensity gravitational fields. Given the same gravitational field,  
	optimal global registration is guaranteed.

	Whereas the down sampling scheme increases the scalability of our method to larger point clouds, it is unable to achieve perfect global minimum through the MPE method alone. Therefore, once the two point clouds are registered with the MPE method, we apply the iterative closest point algorithm to achieve a fast descent to the global optimal registration in the local convex function. Note that such a course to fine approach is applied by all registration methods. However, these methods almost always rely on feature descriptors. On the other hand, the proposed framework for global minimum approximation, as showed in Figure \ref{fig:MPL}, allows the point cloud registration procedure to cross local minima and directly approach the local convex optimization function under the natural force, without the need for feature descriptors.

	\section{Experiments}
	We implemented the proposed method on a PC with Intel i5 3.4Hz processor and designed a series of experiments to test its performance on different data sets. The evaluation metrics are time complexity and estimation error. 
	First, we verify the theoretical properties of our method using the Stanford 3D scanning repository which contains information such as transformation, overlap ratio and ground truth. Next, we compare the proposed method with existing algorithms.
	Finally, we discuss the influence of different hyper-parameter settings on our method. Specifically, we compare the time complexity of the proposed MPE method and its extension to MPL.

	\subsection{Experiments on Synthetic Data}
	In this section, we demonstrate the robustness and convergence properties of the proposed method and compare the results with existing algorithms.
	
	\begin{figure}[t!] 
		\center{\includegraphics[width=0.5\textwidth]{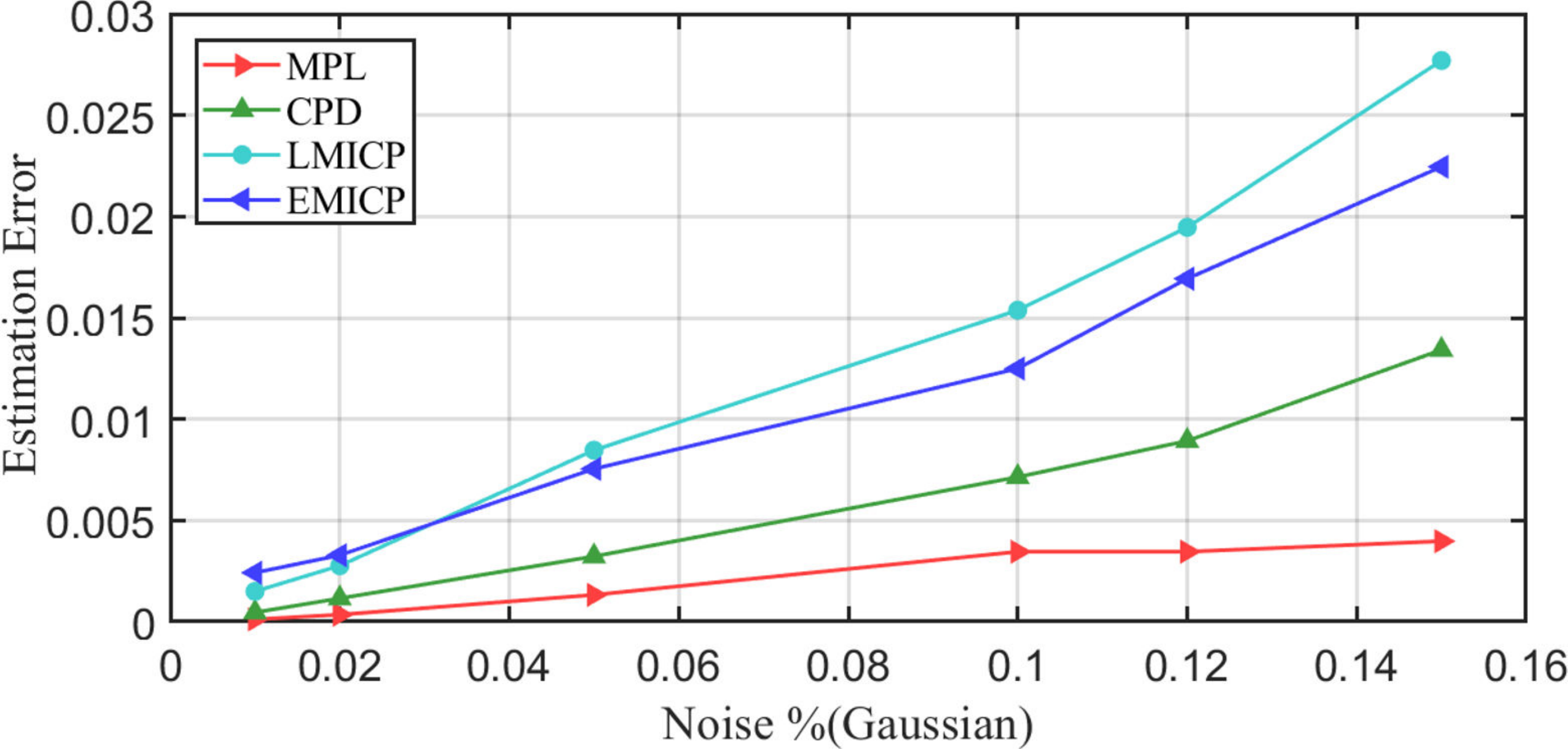}}
		\vspace{-5mm}
		\caption{\label{fig:error_G} Estimation error versus Gaussian noise. The reference set is the Stanford bunny and the template set is randomly transformed and added with 1\% to 15\% Gaussian noise.}
	\end{figure}
	
	\begin{figure}[t!] 
		\center{\includegraphics[width=0.5\textwidth]{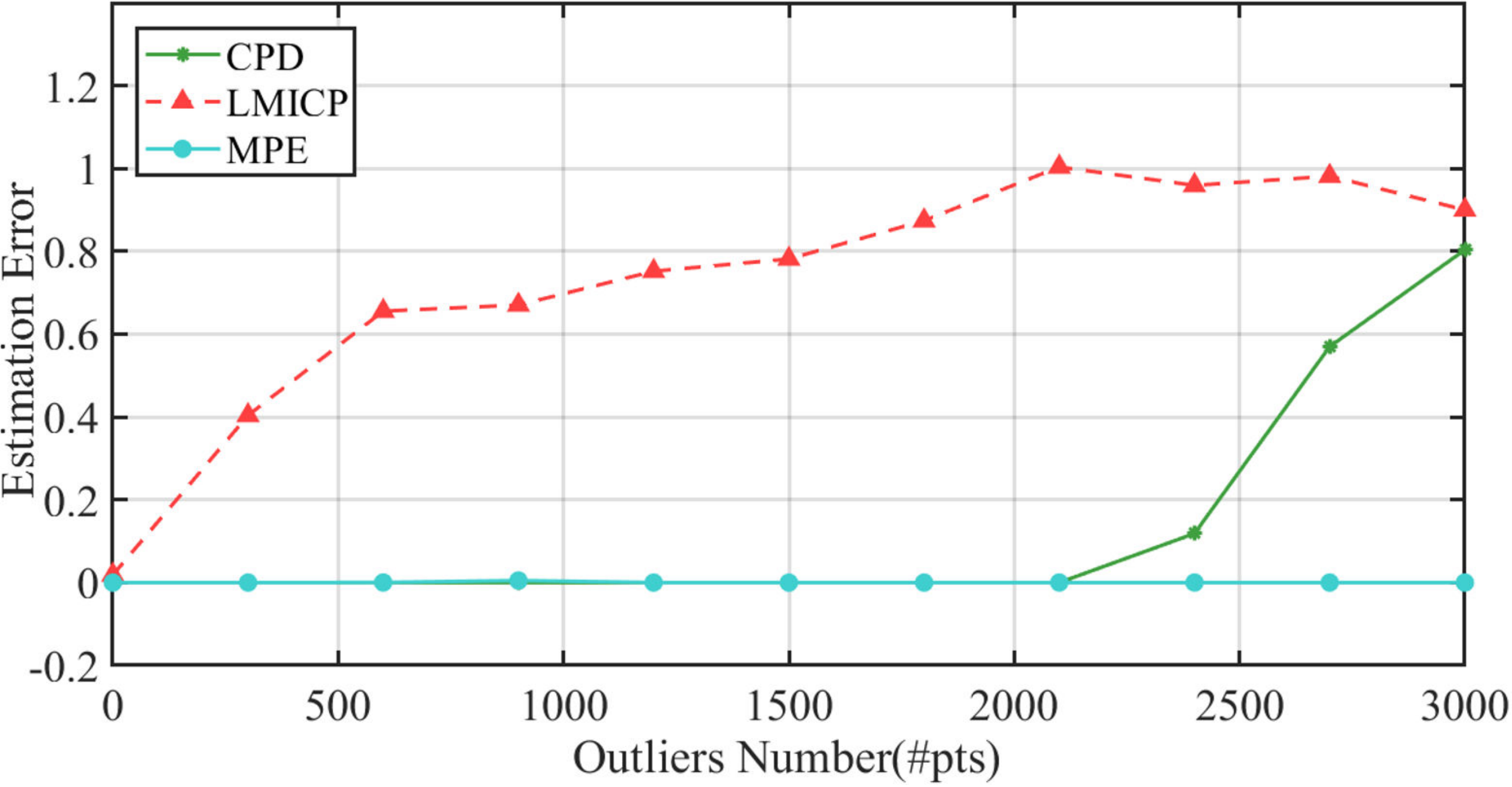}}
		\vspace{-5mm}
		\caption{\label{fig:error_O} Registration time versus outliers. The reference set is the Stanford bunny, and the template set is randomly transformed and added with 0 to 3000 outliers generated from a uniform model.}
	\end{figure}
	
	\emph{$\textit{1)}$ Robustness of the Proposed Algorithm: } We compare our method to the Iterative Closest Point (ICP)~\cite{icp1}, Coherent Point Drift (CPD)~\cite{CPD}, LM-ICP (Levenberg-Marquart ICP)~\cite{OT1} and EM-ICP (Expectation Maximization ICP)~\cite{EMICP} algorithms using the Stanford bunny point cloud data. Specifically, the bunny\_zipper\_res3 (1889 points) was chosen for this experiment. We gave a random translation and rotation (angle $\phi\in [0,\frac{\pi}{2}]$ ) and then introduced Gaussian noise uniformly distributed and added in to template set Y. The Gaussian noise was set from 1\% to 15\% per unit length.

	Figure~\ref{fig:error_G} shows that the proposed MPL algorithm outperforms the CPD~\cite{CPD}, LMICP~\cite{OT1} and EMICP~\cite{EMICP}. We can see that with increasing levels of noise, the registration error of our method remains the lowest and the slope of the curve also remains relatively small. A visual illustration of the registration by our algorithm in the presence of noise is given in Figure~\ref{fig:gaussian_demo}. The blue point set represents the original point cloud, while the red point set is generated with different levels of noise. The translation of every point from the original position is randomly computed by a Gaussian model in 3D space. The standard deviations of the Gaussian distribution are $\sigma=0.01$, $\sigma=0.02$, $\sigma=0.04$, $\sigma=0.08$ and $\sigma=0.16$. As shown in Figure~\ref{fig:gaussian_demo}, our algorithm achieves the lowest error among all algorithms - lower than others by at least $66.50\%$. More interestingly, our method shows less error accumulation for all Gaussian noise levels.

	\begin{figure*}[t!] 
		\center{\includegraphics[width=0.99\textwidth]{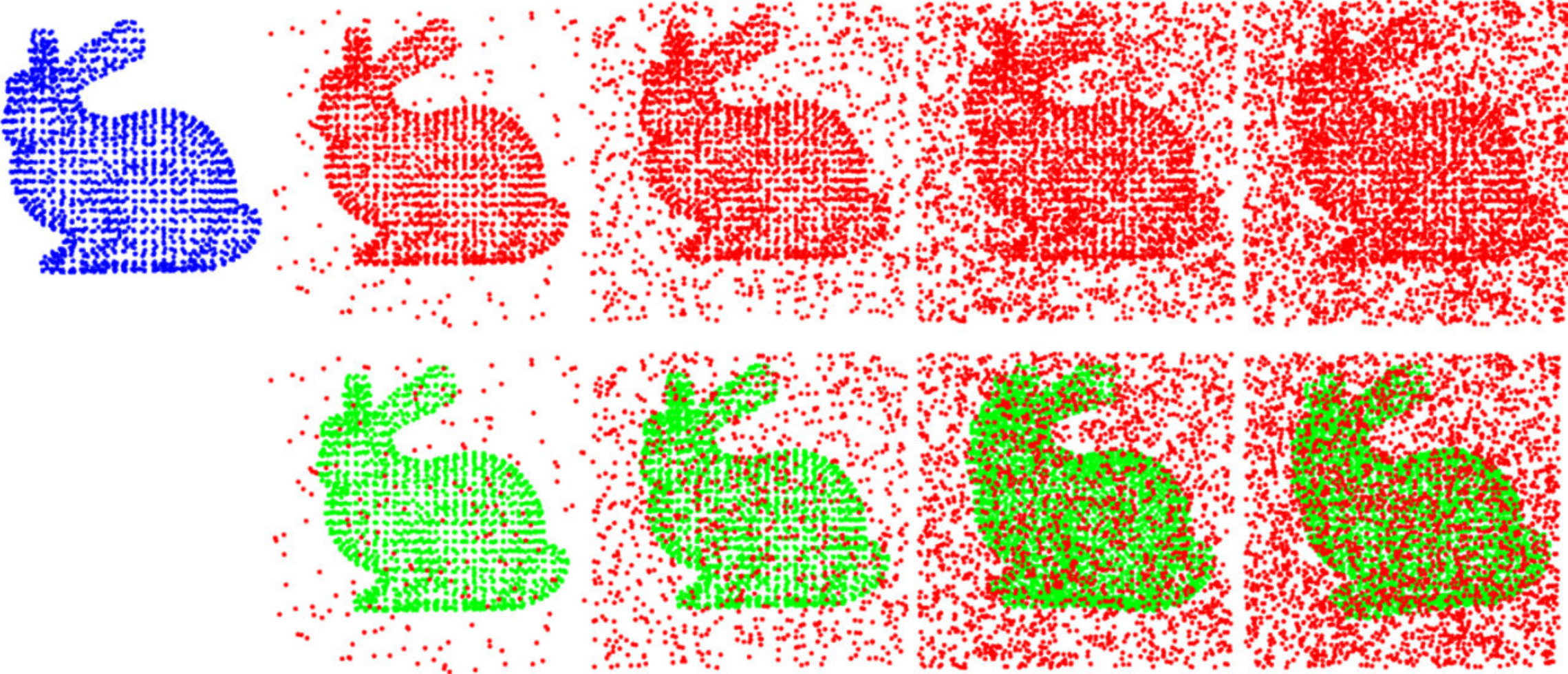}}
		\caption{\label{fig:outlier_demo} Registration results of the proposed method with varying levels of outliers. Fist Row: Reference point cloud (blue) and template point clouds (red) generated from the reference with varying levels of outliers produced by a uniform model in a cube. Second Row: Registration results of the reference point cloud and temple point cloud. For better visualization, we set the registered point pair color to green.}
	\end{figure*}

	Figure~\ref{fig:error_O} compares the estimation error of our method to CPD and LMICP with respect to increasing number of outliers (uniform noise). The outliers ($0 \to 3000$ points or $0 \to 1.5$ times the original points) were generated uniformly in a cube encompassing the point cloud data. The proposed MPL algorithm performs exceptionally well and outperforms CPD and LMICP. Our method shows robustness to a large number of outliers by taking advantage of the global PE minimum to ignore the gravitational field induced by the outliers. The estimation error always stays at a low value approaching zero in this experiment irrespective of the number of outliers. Figure~\ref{fig:outlier_demo} visually demonstrates the registration effectiveness with increasing number of outliers. The original point set is coloured blue, and the red point set is generated by adding random outliers in the data cube following a uniform distribution. The number noise points are $200$, $950$, $2000$ and $2800$. For better visualization, we show the registered model (inliers - with $<0.001$ distance) in green.

	\begin{figure}[t!] 
		\center{\includegraphics[width=0.5\textwidth]{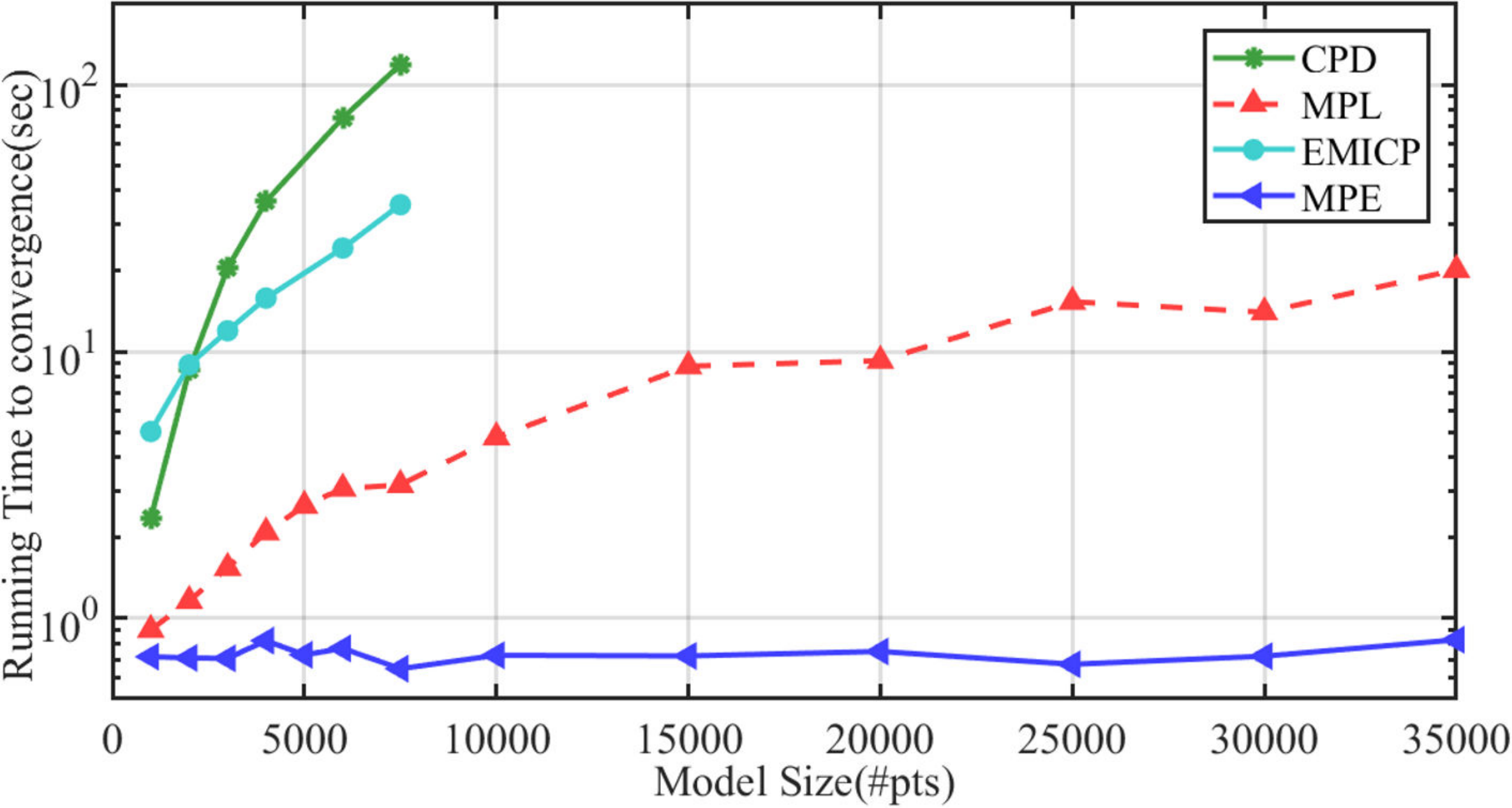}}
		\vspace{-5mm}
		\caption{\label{fig:speed_size} The time cost of the different model sizes is found by applying multiple methods. The left six data points are the comparison between the four methods with points of the model size (from 1000 to 7500 points). The rest of the data points show the performance of our method (MPL and the MPE process of the whole algorithm), while the other method cannot handle it.
		}
	\end{figure}
	
	\emph{$\textit{2)}$ Time to Convergence:} The next experiment is designed to show the time complexity as the model size number is raised. We time the entire running procedure and visualize the results, as in Figure~\ref{fig:speed_size}. The CPD and EMICP are used to compare to our method to illustrate the time complexity performance of our approach. CPD demonstrates the biggest computational cost that rises exponentially. EMICP shows high time complexity at the beginning of the pic, and medium performance in terms of model size increasing. The computational complexity of the two methods above increase exponentially, and the maximum size of the test model is limited within 7500 points. We down-sample the original bunny\_zipper (35947 points) to a list of smaller models that point the number growing from 1000 to 35000. 
	
	A comparison is shown by limiting the model size range within 1000 to 7500. Beyond that, we test the bigger model size on our proposed algorithm individually. Specifically, the blue line represents the time cost of down-sampled MPE procedure in MPL algorithm.The single MPE procedure can run in low time complexity. This causes the centroid capture to need only a few points that maintain the simple shape, and the increasing the model points just make the centroid calculations more approximate to the ground truth, exerting little influence on the global minimum approach. The complete method also shows great performance and low time complexity changes with a lower rate of increase.
	
	\begin{figure}[t!] 
		\center{\includegraphics[width=0.5\textwidth]{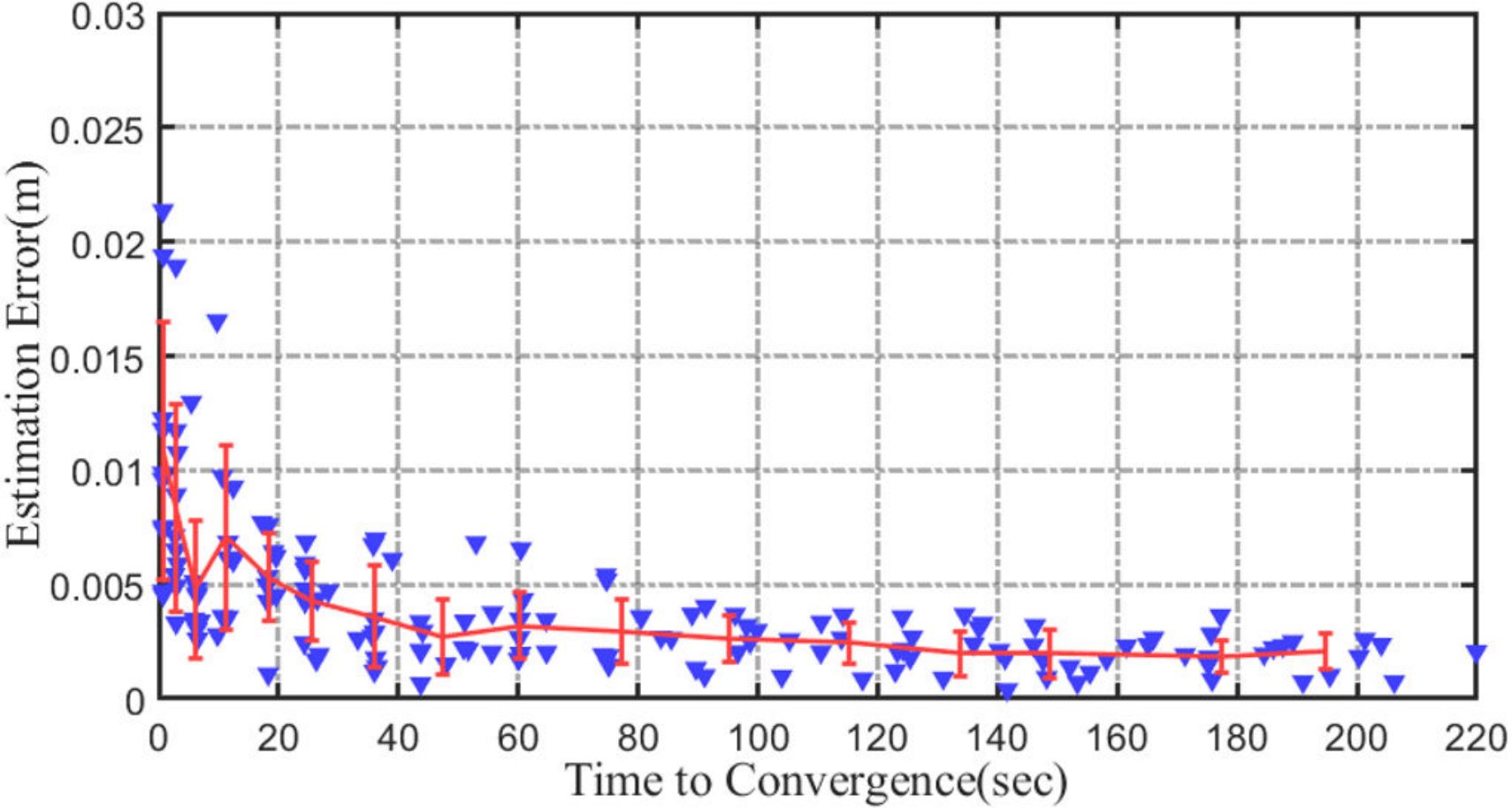}}
		\vspace{-5mm}
		\caption{\label{fig:ET} The results of the estimation error and convergence time with different sampling ratios. The every blue inverted triangle denotes a data point that represents the relation of estimation error and convergence time, under specific sampling ratios. The red line represents their averages and the variance of estimation error with increasing sampling ratio. The metrics are calculated with 160 runs for 16 different sampling ratios}
	\end{figure}
	
	\begin{figure*}[t!] 
		\center{\includegraphics[width=0.99\textwidth]{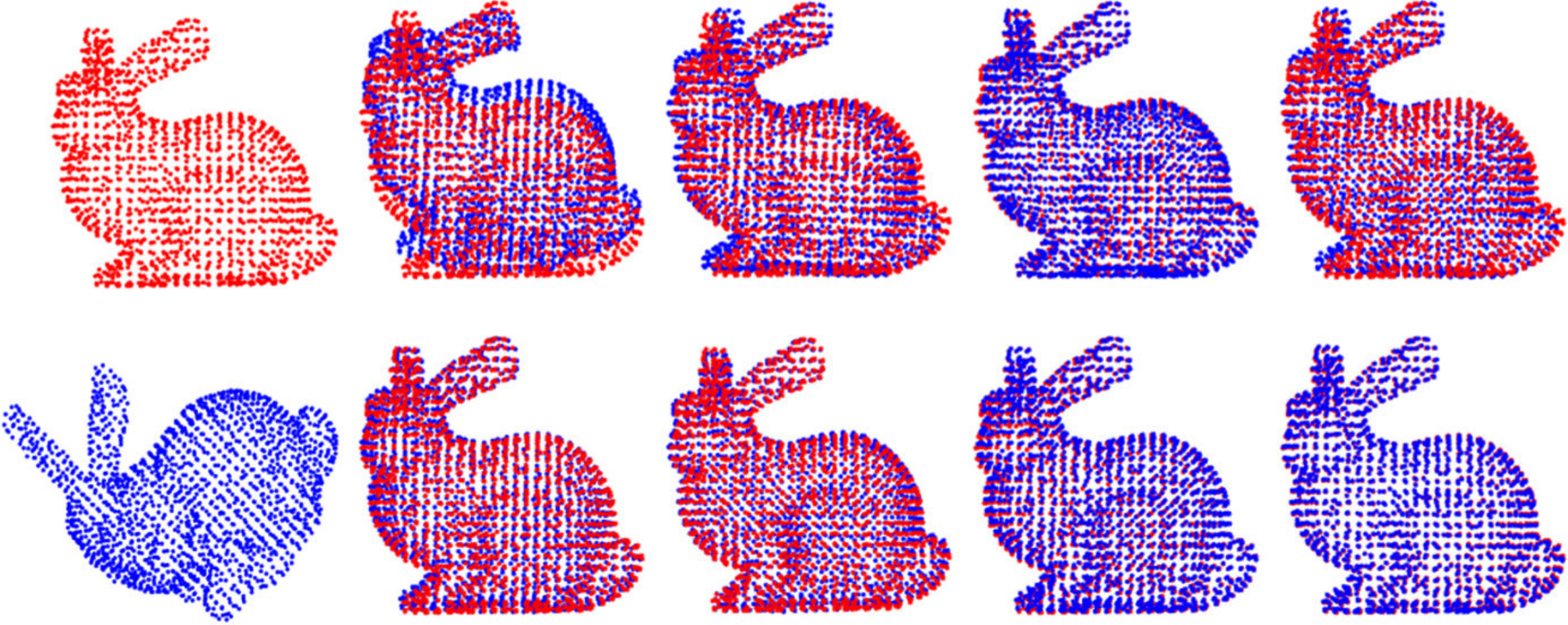}}
		\caption{\label{fig:sampling_D} Registration results of the experiment with varying sampling ratios. The first column: Reference point cloud (top red point set) and temple point cloud (bottom blue point sets). The rest of the figures are the registration results varying with increasing sampling ratios, from top to bottom and left to right.}
	\end{figure*}
	
	To register point clouds with our proposed method, a down-sampling rate is generally required. A higher sampling rate usually not only brings us higher accuracy but also more runtime. Therefore, to maintain the balance between the down-sampling size of MPE computation and algorithm accuracy, we designed an independent experiment to obtain 160 data points of multiple test results from experiments repeated 10 times. In each experiment, we set the down-sampling rate from 5\% to 80\% with an interval of 5\%. Then we tested the time to convergence and the error for each ratio of the down-sampling. Everything mentioned above is visualized in Figure~\ref{fig:ET}. The blue inverted triangle points are the original test data point, whereas the red line represents the time average of each down-sampling and its corresponding standard deviation. Obviously, as the sampling ratio increases, the estimation error becomes smaller and more stable. The significant point is that the error line becomes smoother when the sampling ratio value is over 0.25. The experiment also shows that another threshold is the 0.15 sampling ratio. Once the value is bigger that 0.15, our algorithm can ensure a global optimal more frequently. More detail is showed in Table~\ref{tab:ET}, and the two threshold point data are thickened. The more intuitive demonstration of the sampling ratio change is also visualized in Figure~\ref{fig:sampling_D}. The first column shows the original point cloud, in which the red point set represents the temple set Y and the blue point set is one that has been given a random translation and rotation. Except for this column, the next eight images follow a rule that sampling rate increases progressively from left to right and from top to bottom. 
	
	\begin{table}[htbp]
		\centering
		\caption{\label{tab:MPEL}The Time Cost of Minimum PE Registration For The Naive MPE Theory and Two Down-Sampling Rates of The MPL Algorithm}
		\begin{tabular}{c c c c}
			\toprule
			N$\times$M & MPE & MPL (200) & MPL (100)\\
			\midrule
			453$\times$3	& 16.4s & 3.3s & 0.9s\\
			1889$\times$3	& 5.1m & 3.9s & 1.5s\\ 
			8171$\times$3	& 1.6hr & 7.2s & 4.8s\\ 
			35947$\times$3	& 30.6hr & 20.2s & 18.4s\\ 
			\bottomrule
		\end{tabular}
	\end{table}

	We also tested the performance between the original MPE methods and the modified strategy MPL with a large number of points. Four bunny sets of sizes in Stanford 3D point cloud repository were used: 453$\times$3, 1889$\times$3, 8171$\times$3 and 35947$\times$3. For the down-sampling parameters, we tested the two most common values used in our algorithm: 100 points and 200 points. Table~\ref{tab:MPEL} shows the registration time with MPE and MPL. The MPL algorithm was significantly faster and showed a low registration time growth, no matter the number of sampling points chosen. Furthermore, through comparing the two MPLs (MPL 200 and MPL 100), we noted that the MPE procedure of the fewer points was less time-consuming. Therefore, we concluded that the PE theory is suitable for coarse registration with low time cost to fulfill a more effective point cloud registration.
	
	\subsection{Experiment on real data}
	\emph{$\textit{1)}$ Performance Comparison on UWA Dataset:} We firstly demonstrate the performance of our proposed registration algorithm on the UMA 3D Modeling dataset~\cite{mian1}~\cite{mian2}. These dataset is acquired with a Minolta Vivid 910 scanner and contains several images scanned from four objects. We fulfill the registration of different views by processing the raw point clouds of four objects. Due to the absence of ground truth, we apply the rotation error $\epsilon_r$ and translation error $\epsilon_t$ to measure the performance of registration result. The rotation error $\epsilon^{ij}_r$ and translation error $\epsilon^{ij}_t$ are calculated as following:
	\begin{equation}\label{eq11}  \epsilon^{ij}_r=\frac{180}{\pi}*arcos(\frac{trace(\textbf{R}_{GT}^{ij}*(\textbf{R}_{E}^{ij})^{-1})-1}{2})\end{equation}
	\begin{equation}\label{eq12}  \epsilon^{ij}_t=\|\textbf{T}_{GT}^{ij}-\textbf{T}_{E}^{ij} \| \end{equation}
	Where the $\textbf{R}_{GT}^{ij}$, $\textbf{T}_{GT}^{ij}$ represents the ground truth rotation and translation between the i-th and j-th point sets. The point clouds are coarse aligned manually and then refined with ICP algorithm for the ground truth generation. The $\textbf{R}_{E}^{ij}$, $\textbf{T}_{E}^{ij}$ denotes the estimation rotation and translation respectively. 
	
\begin{figure}[t!] 
		\center{\includegraphics[width=0.5\textwidth]{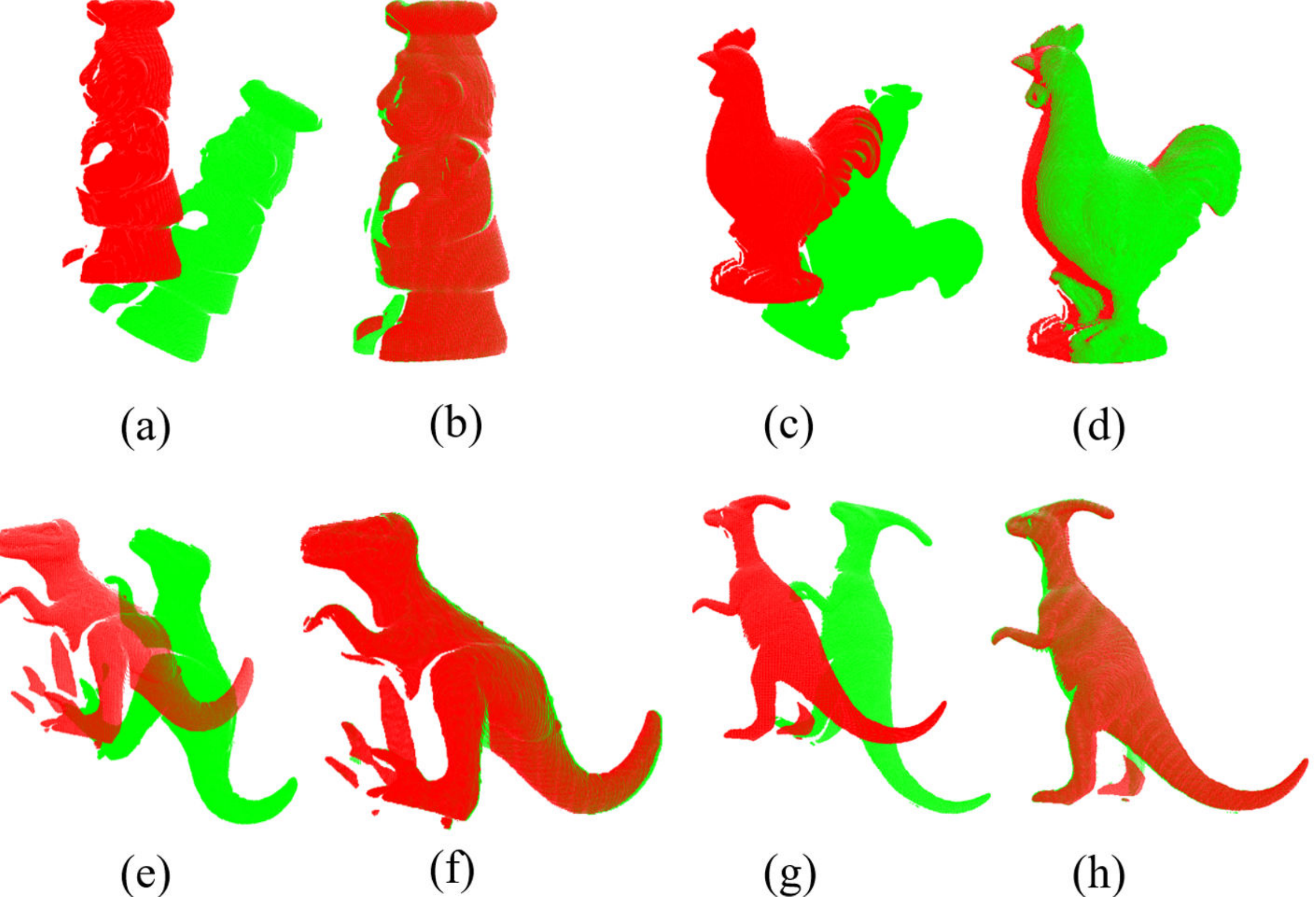}}
		\vspace{-4mm}
		\caption{\label{fig:UWA} The demonstration of registration with our method on the UWA DataSet. Column 1,3 denote the initialization and column 2,4 represent the aligned images. The red images denote the reference sets and the green images represent the template sets transformed with a rotation and translation.}
	\end{figure}

	We perform the comparison of rotation error in Table~\ref{tab:UWAe1} with different methods including Rotational Projection Statistics (RoPs)~\cite{algo2}, Trimmed-icp~\cite{Trimmedicp} and the combination RoPs+generalized-icp~\cite{algo1}~\cite{gedicp}. The RoPs and its related RoPs+generalized-icp algorithms are input with points coordinate and mesh to make use of more information for registration. On the contrary, the trimmed-icp and our proposed method are only input with point coordinate. Furthermore, Our proposed method still achieve the best performance of registration results. Due to the difference of mesh resolution, we report in Table~\ref{tab:UWAe2} the translation error comparison with the trimmed-icp.The demonstration of the registration results on UWA dataset is showed in Figure~\ref{fig:UWA}. The first and third columns show the registration initialization. while the others represent the aligned results. In this experiment, we compare the performance our proposed method with several methods on UWA dataset. The results show that our algorithm achieves the best performance even input with fewer data information.

	\begin{table*}[htbp]
		\centering
		\caption{\label{tab:UWAe1}Rotation Error (deg) of Registration Results of Four Individual Objects on UWA Dataset}
		\begin{tabular}{c c c c c c}
			\toprule
			& 	Input Data	&Chef	&	Chicken	&	Parasaurolophus	& T-Rex\\
			\midrule
			RoPs~\cite{algo2}	&	xyz+Mesh&	2.2117	&	1.0075	&	1.0634	&	1.3722	\\
			RoPs+generalized-ICP ~\cite{algo1}~\cite{gedicp}	&	xyz+Mesh&	0.2712	&	0.3900	&	0.1771	&	0.3758	\\
			Trimmed-icp~\cite{Trimmedicp}	&	xyz		&	5.3751	&	5.0682	&	1.9188	&	1.6655\\
			Ours	&	xyz		&	\textbf{0.2651}	&	\textbf{0.3865}	&	\textbf{0.1527}	&	\textbf{0.3230}\\
			\bottomrule
		\end{tabular}
	\end{table*}
	
	\begin{table}[htbp]
		\centering
		\caption{\label{tab:UWAe2}Translation Error of Registration Results of Four Individual Objects on UWA Dataset. Evaluation Metric is the $L_2$ Distance.}
		\begin{tabular}{c c c}
			\toprule
			&Trimmed-icp	&	Ours\\
			\midrule
			Chef		&		6.9117	&	\textbf{0.5775}\\
			Chicken		&		0.8773	&	\textbf{0.4753}\\
			Parasaurolophus	&		7.1050	&	\textbf{0.3031}\\
			T-Rex			&	3.9467	&	\textbf{0.4006}\\
			\bottomrule
		\end{tabular}
	\end{table}
	
	\emph{$\textit{2)}$ Specific Blade Point Cloud Registration:} This experiment shows the ability of dealing with specific point clouds. Due to the local features of the neighboring points being extremely similar, extracting the feature of the surface object showing a complex special-shaped surface (example, aero-engine blade) is difficult. In this situation, the traditional descriptor-based method~\cite{icp1}~\cite{lei2017fast}~\cite{FPFH} is not useful. We demonstrate the comparison of the performance between our proposed algorithm and the traditional descriptor-based method, FPFH. The point cloud data is scanned and processed from the aero-engine blade. 
	
\begin{figure}[t!] 
		\center{\includegraphics[width=0.5\textwidth]{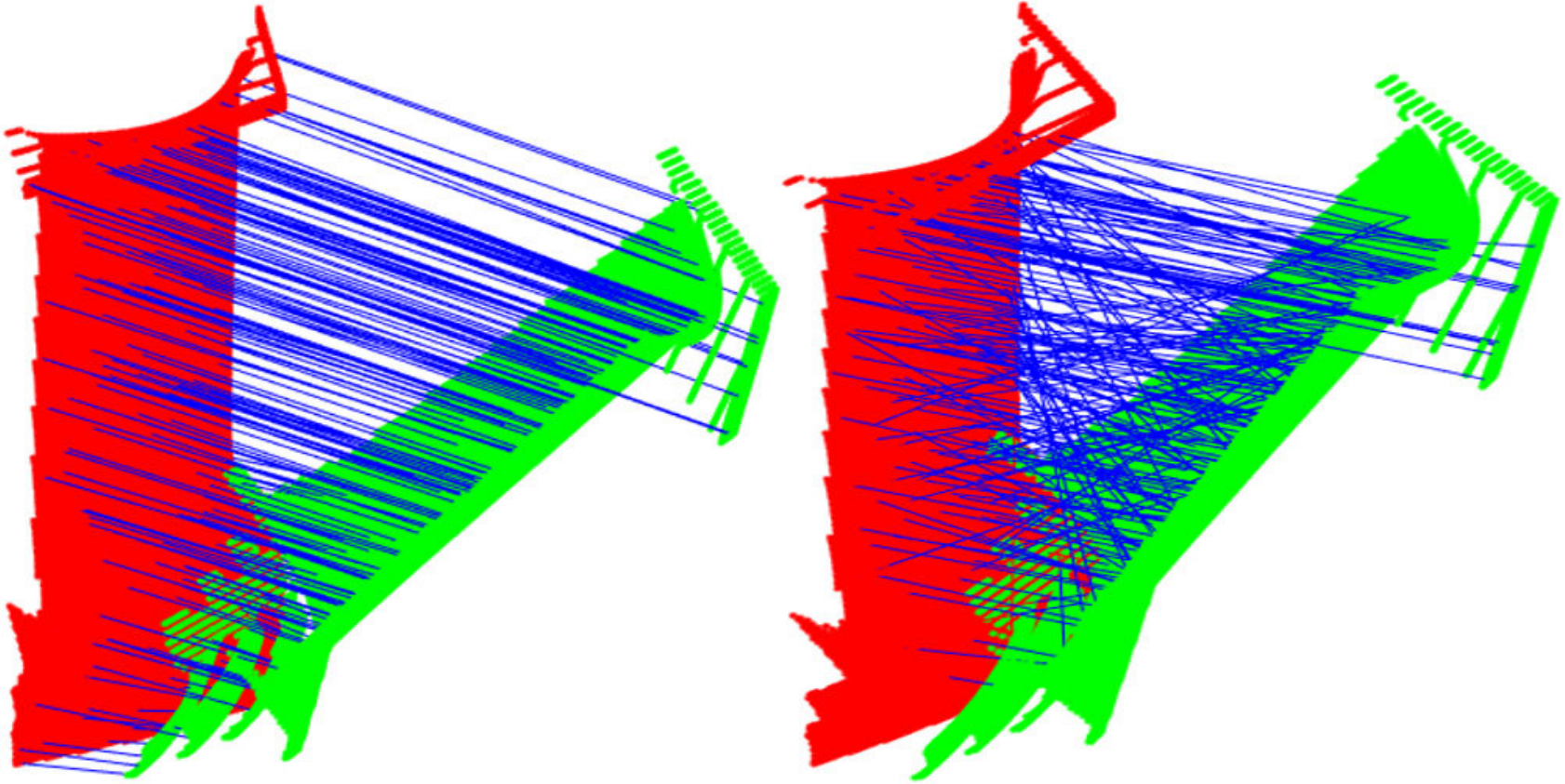}}
		\vspace{-5mm}
		\caption{\label{fig:leaf_demo} The comparison of the registration results between our proposed method and the FPFH (the descriptor-based method). Data: Reference point cloud (red point set) and temple point cloud (green point sets). The blue lines connects the registered point pair. For better comparison, the conception that the point pair registered by our method (left figure) is assumed as a correspondence. And the FPFH correspondence point pair is the right-side figure.
		}
	\end{figure}

	Figure~\ref{fig:leaf_demo} shows the effectiveness of the registration between the two methods. The original data is showed as the red point set and the green set is translated and rotated from the original point cloud. Furthermore, the blue line connects the point pair of the inliers. The right image shows the registration point pair of the FPFH method using feature extraction, and ours is showed on the left. For a more intuitive visualization, the point pairs of the inliers are partially displayed. With our method, the point pair is considered as the correspondence, especially when the distance of it is less than a certain threshold. However, when dealing with the blade surface point cloud, the FPFH method tends to fail. Moreover, the correctness matching rate is less than 15\% with the whole blade, whereas the matching accuracy of the surface part is less than 4\%. The precision of the matched is defined as: \begin{equation}\label{eq10} \text{precision}=\frac{\text{The number of correct matches}}{\text{The number of ground truth matches}} \end{equation}
	Our proposed method registers the point cloud without relying on a feature-based theory. It is significant to deal with complex, special-shaped surfaces that lack the texture of information, and the method has proven its superiority.
	
	\begin{figure*}[t!] 
		\center{\includegraphics[width=0.99\textwidth]{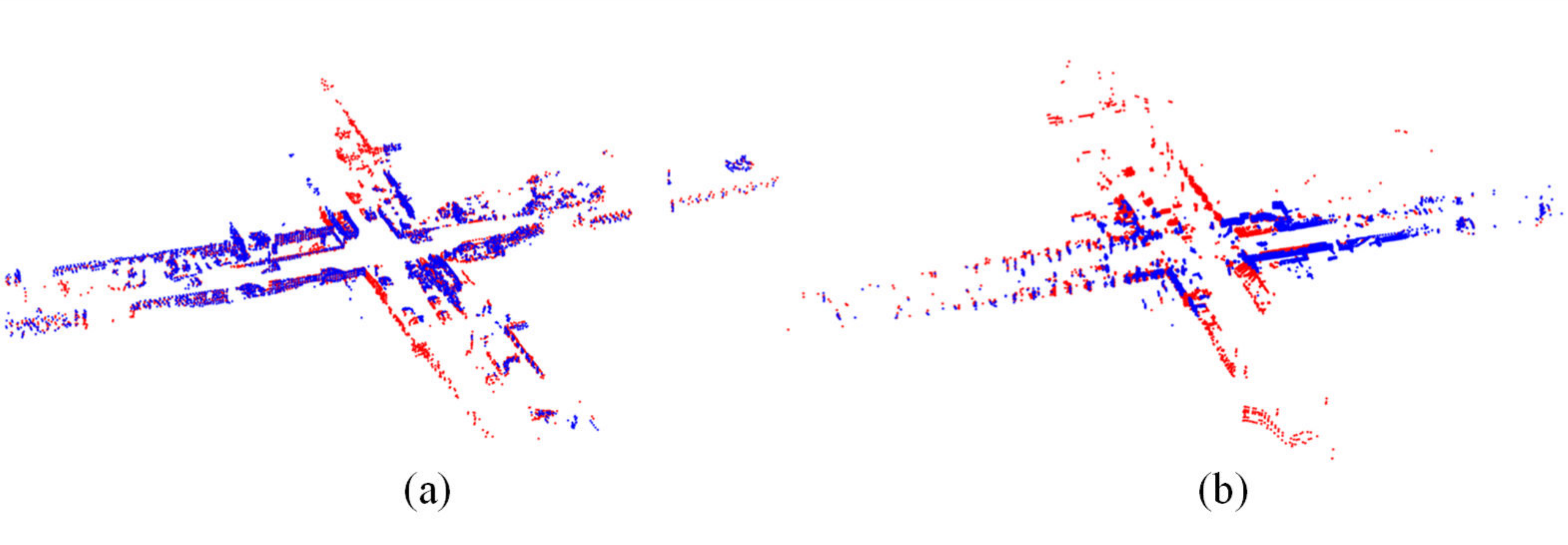}}
		\caption{\label{fig:real_street} The demonstration of real the street data registration results. Four scenes in~\cite{dataset} are selected to perform the ability of large scale shape feature extraction. (a) shows the high overlap urban lidar street data registration. On the contrary, (b) is only with few overlap on cross shape feature. The red,blue point sets denote the reference sets and the template sets respectively.}
	\end{figure*}
	
	\emph{$\textit{3)}$ City lidar Point Cloud Registration: } Finally, we conduct a test on city lidar data using our proposed method of urban registration. The \textbf{CITY} scans were collected by Alastair Quadros in 2013, and contain a variety of common urban road objects with wide-baselines, which is quite difficult to reconstruct. The testing point clouds were extracted from Sydney's urban objects dataset~\cite{dataset}. We selected several scenes of long streets to demonstrate the registration with large scale global shape feature extraction, whereas the traditional methods (such as icp, FPFH, and etc.) can hardly achieve it.
	
	Here, we register the urban road point cloud and shown in Figure~\ref{fig:real_street}. Figure~\ref{fig:real_street}(a) shows a high overlap condition (the shape overlaps more than 80\%) of the real road lidar point cloud registration and displays a fine performance. Another situation that contains a lower overlap and less information retained except, basically, the shape, is shown in Figure~\ref{fig:real_street}(b). It is clear that our method has a fine performance in urban road point cloud data application and can easily register the global feature, for instance, the two street and a cross in common urban scenes. 
	
	Real city lidar data contains a wide scale sparse point cloud, while the two scenes point clouds hardly overlap. The ground truth of the point pair are not known, and the truth correspondence is biased due to the wide scale scan procedure. To deal with the different situations, since the traditional method relying on the high accuracy of the source point cloud tends to fail, our method introduces the different weights of different parts generated by the density of the points at different locations. The proposed method can extract the global shape feature to fulfill a more accurate registration, which is the most important application of our theory.The dense clusters present more information, and our method distributes more weight to the kind of cluster to obtain a fine registration result. The results show that our method performs great within a large scale city lidar point cloud; the proposed method still stays efficient and accurate.

	\section{Conclusion}
	In this paper, we proposed a novel, globally optimal theory for point cloud registration in 3D, which was achieved by introducing a specific physical system to approach the MPE of the point clouds. We also proposed a new $P_2$ criterion for the over-fitting problem. Our method of gathering the point clouds through rotation torque and gravitational vector fulfilled the globally optimal registration efficiently. The proposed approach was able to align point clouds within large quantities of Gaussian and uniform noise, while running with low time complexities. We showed a fine performance with a comparison between our proposed method and some other approaches. Besides, we discussed the drawbacks of traditional descriptor-based methods and explored a new path to offer a new solution. The proposed approach was also able to register a variety of common urban road objects.

	\appendices
	%

	\section*{Acknowledgment}
	This work was supported in part by National Natural Science Foundation of China under Grant 61573134, Grant 61733004 and in part by the Australian Research Council under Grant DP190102443.

	\ifCLASSOPTIONcaptionsoff
	\newpage
	\fi
	
	\bibliographystyle{IEEEtran}
	\bibliography{Reference}
	


	
\end{document}